\documentclass[pdflatex,sn-basic]{sn-jnl}

\usepackage{graphicx}
\usepackage{multirow}
\usepackage{amsmath,amssymb,amsfonts}
\usepackage{amsthm}
\usepackage{mathtools}
\usepackage[title]{appendix}
\usepackage{xcolor}
\usepackage{textcomp}
\usepackage{booktabs}
\usepackage{siunitx}
\usepackage{enumitem}
\usepackage{algorithm}
\usepackage{algorithmicx}
\usepackage{algpseudocode}
\usepackage{xspace}

\usepackage[capitalise,noabbrev]{cleveref}   

\crefname{equation}{Eq.}{Eqs.}
\Crefname{equation}{Eq.}{Eqs.}

\theoremstyle{thmstyleone}

\theoremstyle{thmstyletwo}

\theoremstyle{thmstylethree}

\newif\ifdraft\drafttrue
\ifdraft
  \newcommand{\todo}[1]{\textcolor{red}{\textbf{[TODO: #1]}}}
  \newcommand{\note}[2]{\textcolor{blue}{\textbf{[#1: #2]}}}
\else
  \newcommand{\todo}[1]{}
  \newcommand{\note}[2]{}
\fi

\newcommand{\R}{\mathbb{R}}

\newcommand{\set}[1]{\mathcal{#1}}
\newcommand{\vect}[1]{\mathbf{#1}}
\newcommand{\mat}[1]{\mathbf{#1}}
\newcommand{\norm}[1]{\left\lVert#1\right\rVert}

\DeclareMathOperator*{\argmin}{arg\,min}

\newcommand{\method}{DC-ViT\,v2\xspace}
\newcommand{\img}{\vect{x}}            

\newcommand{\nchan}{C}                  
\newcommand{\npatch}{N}                 
\newcommand{\ntok}{N}                   
\newcommand{\dmodel}{D}                 
\newcommand{\nlayer}{L}                 

\newcommand{\tokens}{\vect{z}}            
\newcommand{\lay}{\ell}                 
\newcommand{\mixset}{\set{S}}           
\newcommand{\amix}{\alpha}              
\newcommand{\Wo}{\mat{W}_{\!O}}         
\DeclareMathOperator{\Attn}{Attn}
\DeclareMathOperator{\MSA}{MSA}
\DeclareMathOperator{\DSA}{DSA}
\DeclareMathOperator{\MLP}{MLP}
\DeclareMathOperator{\softmax}{softmax}
\newcommand{\Attnsp}{\Attn_{\mathrm{sp}}}   
\newcommand{\Attnch}{\Attn_{\mathrm{ch}}}   
\newcommand{\gsp}{g_{\mathrm{sp}}}          
\newcommand{\gch}{g_{\mathrm{ch}}}          
\newcommand{\amixhat}{\hat{\alpha}}          
\newcommand{\chanref}{c_{\mathrm{ref}}}      
\newcommand{\chanq}{c_{\mathrm{q}}}          
\newcommand{\loc}{\mathrm{loc}}              

\newcommand{\best}[1]{\textbf{#1}}

\title[DC-ViT for MCI data]{Extending Decoupled Attention to Dense Prediction and Masked Training for Multi-Channel Images}

\author{\fnm{Umar} \sur{Marikkar}}
\author{\fnm{Sameed} \sur{Husain}}
\author{\fnm{Muhammad} \sur{Awais}}
\author{\fnm{Sara} \sur{Atito}}

\affil{\orgname{{University of Surrey}}}

\abstract{Multi-Channel imaging (MCI) data differs fundamentally from natural images, as each channel records a semantically distinct signal rather than a colour band. To adapt vision encoders to MCI data, Multi-Channel Vision Transformers (MC-ViTs) tokenize each channel independently and concatenate the resulting tokens into one sequence, and the channel count is no longer fixed by the architecture. Self-attention is then computed across all channel-patch tokens with no restriction on which channels attend to which, which dilutes the features of individual channels. The Decoupled Vision Transformer (DC-ViT) regulates this by separating updates computed within a channel from updates computed across channels, and by forming a representation per channel before the channels are combined. Its formulation, however, pairs tokens by spatial position, and thus requires the same visible tokens in every channel. Correspondence under independent per-channel masking is recovered by solving a linear assignment between the retained patches of each channel, which allows decoupled attention to be combined with current masked multi-channel training in its standard configuration rather than a restricted one. Across three classification and three segmentation benchmarks spanning fluorescence microscopy, imaging mass cytometry and satellite imaging, including dense prediction at high channel counts, the resulting formulation outperforms the strongest MC-ViT baseline.}

\keywords{multi-channel images, channel-agnostic
models, transfer learning}

\begin{document}

\maketitle

\section{Introduction}
\label{sec:intro}

A multi-channel image (MCI) records several measurements of the same scene, where each channel corresponds to a separate physical or biological quantity. In immunofluorescence microscopy, each channel is a stain that marks a specific subcellular structure \citep{thul2017subcellular}. In imaging mass cytometry, each channel is the abundance of one labelled antibody \citep{jackson2020single}. In satellite imaging, each channel is a spectral band with distinct reflectance properties \citep{sumbul2019bigearthnet}. A channel therefore carries meaning on its own, and two channels of the same image are not interchangeable in the way the red and green channels of a photograph are. Vision encoders for general computer vision are not built for this. Standard vision encoders \citep{he2016deep,dosovitskiy2021image} are trained for a fixed number of input channels, and a model trained on one set of channels cannot be applied to another. However in MCI data, the channel counts and the physical quantity that each channel measures all change between studies, since they follow from different staining or sensory protocols. When re-using encoders for MCI data, the common solution is to pre-train channel-agnostic encoders and use them as frozen feature extractors \citep{bourriez2024chadavit,marikkar2025c3r}, but inter-channel relationships which can be beneficial for the downstream tasks are not learned in this setting. Therefore, Multi-Channel Vision Transformers (MC-ViTs) \citep{bao2024channelvit,pham2024dichavit} have been proposed as fine-tuned encoders for downstream tasks. 

Following ViT nomenclature, MC-ViTs remove the fixed-channel constraint by treating each channel as its own set of tokens, and an image with any number of channels becomes one token sequence. Attention then runs over that sequence, and every token may attend to every other token irrespective of the channel it came from. Existing methods preserve channel identity by tagging tokens with the channel they belong to \citep{bao2024channelvit}, or by adding losses that reward diversity between channel features \citep{pham2024dichavit}, but the attention operation itself stays unconstrained. How much information moves between channels is thus decided implicitly during training, and not explicitly regulated. We find that neither unconstrained mixing nor full isolation between channels is appropriate for MCI data. When mixing is unconstrained, channel representations converge, and the encoder no longer preserves what each channel measures. When channels are fully isolated, useful context is lost, since some channels are interpretable relative to others. For example, in the HPA dataset \citep{thul2017subcellular}, a protein stain is read against the nuclear stain that locates it. What is required is control over how much information passes between channels, and over where in the network it passes. This is currently formulated via the Decoupled Vision Transformer (DC-ViT) \citep{marikkar2026dcvit}, which combines decoupled attention and channel-aware aggregation. Here, decoupled attention provides the information control by separating updates computed within a channel from updates computed across channels. Channel-aware aggregation carries the same control into pooling, where a representation is formed per channel before the channels are combined.

The existing implementation of DC-ViT requires the same visible tokens in every channel, which is incompatible with the independent per-channel masking used in current masked multi-channel training \citep{pham2025chamaevit}, and consequently restricts comparison to a weaker configuration of it. Rather than requiring identical visible tokens, correspondence under independent masking is recovered by pairing the retained patches of two channels so that the total distance between partners on the grid is smallest. We then evaluate the result across classification and segmentation on six benchmarks spanning fluorescence microscopy, imaging mass cytometry and satellite imaging, at channel counts up to 40. The aim is a formulation of decoupled attention that applies across MCI data and tasks.

In summary, this paper proposes the following contributions:
\begin{itemize}[leftmargin=*,itemsep=2pt]
  \item We show that decoupled attention extends to dense prediction, and that it holds at high channel counts on imaging mass cytometry, where joint attention over all channel-patch tokens is most costly.
  \item We recover token correspondence under independent per-channel masking by a selection algorithm on the patch grid, which allows decoupled attention to be combined with current masked multi-channel training in its standard configuration.
\end{itemize}
\section{Related Work}
\label{sec:related}

\subsection{Channel semantics in multi-channel imaging}

In general computer vision, the three channels of an RGB image are measurements of the same visible-light signal. Multi-channel imaging does not behave this way. In immunofluorescence microscopy a channel is one stain, and the panel is chosen per study, meaning channels refer to different markers in different datasets \citep{thul2017subcellular,chandrasekaran2023jump,viana2023integrated}. Some markers appear in every image and locate subcellular structures, while others vary with the condition being tested, and the second group is often read against the first \citep{marikkar2025c3r}. Remote sensing has the same property. The bands of So2Sat LCZ42 span visible, near-infrared and short-wave infrared ranges, and each responds to a different property of the surface \citep{zhu2019so2sat}. Similarly, imaging mass cytometry extends marker counts to tens of antibody channels in a single acquisition. In all of these the channel axis carries information that the spatial axis does not, and this information is often distinct unlike RGB channels in natural images. 

\subsection{Multi-channel vision transformers}

MC-ViTs are the existing implementation of vision encoders, specifically ViTs, to MCI data. Here, a shared patch embedding is applied to each channel separately, and the resulting tokens are concatenated into one sequence with positional encodings that align them across channels. ChAdaViT \citep{bourriez2024chadavit} and ChannelViT \citep{bao2024channelvit} add a learned embedding identifying the channel a token came from, and ChannelViT samples channel subsets during training so that inference does not require the full set. DiChaViT \citep{pham2024dichavit} shows that tokens still homogenise across channels, and adds sampling and regularisation terms that reward diversity between channel features. ChA-MAEViT \citep{pham2025chamaevit} combines these ideas with masked autoencoding \citep{he2022mae}, using masking that varies over both channels and patches, memory tokens shared across the sequence, and a channel-aware decoder. A separate line pretrains on individual channels and combines them only when fine-tuning \citep{lian2025isolated}. The above works on MC-ViTs have gradually introduced methods to increase inter-channel diversity. However, what none of these change is the attention itself, which is computed over the joint sequence of all channel-patch tokens. Inter-channel interaction is instead regulated through tokens, sampling and losses, rather than through which tokens are permitted to interact. 

In contrast, DC-ViT \citep{marikkar2026dcvit} applies this regulation architecturally. Here, attention is computed twice within a block, once over the tokens of a given channel and once over the channels at a given spatial position, and the two results are blended by a learned scalar. Further, channel attention is applied at only a subset of blocks, hence earlier blocks build channel representations independently before any mixing occurs. The same separation carries into pooling, where the final representation is formed within each channel before the channels are combined. However, DC-ViT is evaluated on classification alone, and its current formulation requires strict spatial correspondence between channels.
\section{Methodology}
\label{sec:method}

\subsection{Preliminaries}
\label{sec:prelim}

\paragraph{The MC-ViT encoding protocol.}
Let a multi-channel image be $\img \in \R^{\nchan \times H \times W}$, where $\nchan$ is the number of channels and $H, W$ are the spatial dimensions. MC-ViT applies a shared patch embedding to each channel separately, so that a channel contributes $\npatch$ tokens and the image is represented by $\nchan \npatch$ tokens in total. Channel and spatial positional embeddings are added, so that a token carries both the channel it came from and its location within that channel. Auxiliary tokens such as $\mathtt{cls}$ tokens and memory tokens/registers are then prepended to the sequence~\citep{pham2025chamaevit}. For brevity, we exclude the auxiliary tokens from the following equations. Flattening the channel and spatial axes together yields a feature set of $\tokens_\lay \in \R^{\nchan \npatch \times \dmodel}$ at block $\lay$, and each block applies the standard transformer update,
\begin{align}
\label{eq:mcvit-block}
  \tokens_{\lay+1} &= \tokens_\lay + \MLP\!\left(\tokens_\lay + \MSA(\tokens_\lay)\right),
\end{align}

where $\MSA(\tokens_\lay) = \Wo \Attn(\tokens_\lay)$ and $\Attn$ is scaled dot-product attention,
\begin{align}
\label{eq:attn}
  \Attn(\tokens) &= \softmax\!\left(\frac{\vect{q}\vect{k}^{\top}}{\sqrt{\dmodel}}\right)\vect{v},
\end{align}

with $\vect{q}, \vect{k}, \vect{v}$ obtained by linear projection of $\tokens$. The point to note is where the softmax is taken. It runs over the whole flattened sequence, hence a token at one location in one channel may attend to any token at any location in any other channel, and nothing in the operation distinguishes a within-channel interaction from a cross-channel one. Attention over a sequence of length $\nchan\npatch$ is quadratic in that product, and compute cost grows in the order of $(\nchan\npatch)^2$.

\paragraph{Masked image modelling for MCI.}
ChA-MAEViT \citep{pham2025chamaevit} trains an MC-ViT with a reconstruction objective alongside the task loss. Patches are masked independently in each channel at a fixed ratio, which leaves a different visible set in every channel. Whole channels are masked as well, in a number drawn per sample, and the model reconstructs them from the channels that remain. A single decoder, shared across all channels, reconstructs the pixels of the masked patches under an $L_2$ loss on the pixels and an $L_1$ loss in Fourier space. The two objectives are combined as $\mathcal{L} = (1 - \lambda_{\mathrm{rec}}) \mathcal{L}_{\mathrm{task}} + \lambda_{\mathrm{rec}} \mathcal{L}_{\mathrm{rec}}$, with $\lambda_{\mathrm{rec}} = 0.99$, so reconstruction carries most of the weight. Trained this way, MC-ViTs outperform their counterparts trained on the task loss alone across MCI benchmarks. We therefore expect masked training to be the standard setting for MCI going forward, and treat it as the setting a method has to work in.

\paragraph{Decoupled Vision Transformer.}
In DC-ViT \citep{marikkar2026dcvit} tokens are held as $\tokens_\lay \in \R^{\nchan \times \ntok \times \dmodel}$, with the channel axis kept separate without flattening. Attention is then computed twice from a single set of projections. The spatial pass attends within a channel, over that channel's own $\ntok$ tokens,
\begin{align}
\label{eq:attn-sp}
  \Attnsp(\tokens_\lay) &= \left[\Attn(\tokens_{\lay,c})\right]_{c=1}^{\nchan}, \qquad \tokens_{\lay,c} \in \R^{\ntok \times \dmodel},
\end{align}

while the channel pass attends across channels at a fixed spatial location,
\begin{align}
\label{eq:attn-ch}
  \Attnch(\tokens_\lay) &= \left[\Attn(\tokens_{\lay,\cdot,n})\right]_{n=1}^{\ntok}, \qquad \tokens_{\lay,\cdot,n} \in \R^{\nchan \times \dmodel}.
\end{align}

Here the two passes are combined by a learned scalar $\amix_\lay$, and the result replaces $\MSA$ in \Cref{eq:mcvit-block},
\begin{align}
\label{eq:dsa}
  \DSA(\tokens_\lay) &=
  \begin{cases}
    \Wo\!\left((1-\amix_\lay)\,\Attnsp(\tokens_\lay) + \amix_\lay\,\Attnch(\tokens_\lay)\right), & \lay \in \mixset,\\[2pt]
    \Wo\,\Attnsp(\tokens_\lay), & \lay \notin \mixset,
  \end{cases}
\end{align}

where $\mixset \subseteq \{1, \dots, \nlayer\}$ is the set of blocks that carry channel attention. Blocks outside $\mixset$ are hence ordinary transformer blocks applied to each channel independently, and channels exchange nothing there. Because both passes read the same projections, parameter counts are the same, and the attention cost of a decoupled block is $\nchan\npatch^2 + \npatch\nchan^2$ rather than $(\nchan\npatch)^2$.

The final representation follows the same intuition. Rather than pooling from one undifferentiated set of tokens, a pooling function $\gsp$ is applied within each channel and a second function $\gch$ then collapses the channel axis,
\begin{align}
\label{eq:readout}
  \vect{y}_c &= \gsp(\tokens_{\nlayer,c}) \in \R^{\dmodel}, \qquad
  \vect{y} = \gch\!\left(\left[\vect{y}_c\right]_{c=1}^{\nchan}\right) \in \R^{\dmodel}.
\end{align}

Here $\gch$ is a maximum over channels, taken independently for each feature dimension, so that every feature of $\vect{y}$ is read from whichever channel responds most strongly to it. One quantity in this formulation is left open, and it is the subject of \Cref{sec:design}. It is implicit in \Cref{eq:attn-ch}, which pairs tokens by their spatial index $n$, and thus presumes that spatial location $n$ refers to the same spatial location in every channel. This is violated when per-channel random masking in carried out in recent MC-ViT studies \citep{pham2025chamaevit}.

\subsection{\method}
\label{sec:design}

\paragraph{Token correspondence under random masking.}
Current masked multi-channel training draws an independent patch mask for each channel \citep{pham2025chamaevit}, which has been shown to improve downstream performance over masking the same patches in every channel. The channels of a sample no longer expose the same patches, and this has an effect on \cref{eq:attn-ch}, since pairing tokens by spatial index assumes that index $j$ names the same location in every channel. Under independent masks that correspondence has to be established, and we establish it as follows.

We begin with two channels $c_1$ and $c_2$ of the same sample, each carrying its own random patch mask. Let the sets $\set{K}_1$ and $\set{K}_2$ be their retained patches, with $k$ patches in each set. Patches are indexed as $p = 1, \dots, \npatch$, and $\loc(p) \in \R^2$ gives the coordinates at which patch $p$ sits on the grid. Putting the two channels in correspondence means giving every patch of $c_1$ a partner in $c_2$, with no patch being used twice. Such a pairing is a one-to-one map $m'$ from $\set{K}_1$ to $\set{K}_2$, so that the patch of $c_2$ paired with patch $p$ of $c_1$ is $m'(p)$. The optimal mapping $m$ is chosen by minimising the following objective,
\begin{align}
\label{eq:assign}
  m &= \argmin_{m'} \; \sum_{p \in \set{K}_1} \norm{\loc(p) - \loc(m'(p))}^2 ,
\end{align}

where each term of the sum is the squared grid distance between a patch of $c_1$ and the patch of $c_2$ paired with it. This is solved by the Hungarian algorithm \citep{kuhn1955hungarian,jonker1987shortest}. Partnered patches are then placed at the same spatial index in \cref{eq:attn-ch}, so that channel attention compares them.

\begin{figure}[t]
  \centering
  \includegraphics[width=\linewidth]{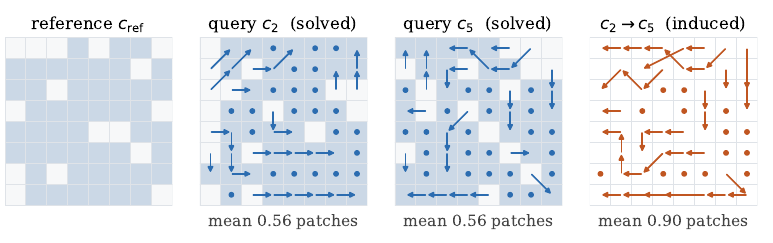}
  \caption{Correspondence under random masking, with $\nchan = 8$ channels on an $8 \times 8$ patch grid at a mask ratio of $0.25$. Shaded cells are retained patches, arrows run from a reference patch to the partner assigned to it by \cref{eq:assign}, and dots mark patches matched to themselves. The rightmost diagram is the induced correspondence between the two queries.}
  \label{fig:masking-example}
\end{figure}

\Cref{eq:assign} pairs two channels, whereas channel attention compares one spatial index across all $\nchan$ channels at once, hence every channel must agree on a single labelling of indices. Solving that jointly is the multi-dimensional assignment problem, which is NP-hard for $\nchan \geq 3$ \citep{karp1972reducibility}, hence no exact solution is available at practical cost. We instead score a set of channels by its distances to one channel alone, rather than by the distances between every pair in it. This is known as a star cost \citep{walteros2014star}, the one channel being the centre of the star. We call it the reference $\chanref$, and treat each remaining channel as a query $\chanq$ to be assigned against it. Under this cost the joint problem separates, and its solution is exactly $\nchan - 1$ applications of \cref{eq:assign}, and two query channels are placed in correspondence only through the reference, as shown in \cref{fig:masking-example}.

Through this however, the reference channel is systematically better aligned than the others, as the correspondence between two query channels is only built implicitly. At $\nchan = 8$ and a mask ratio of $0.25$, a fixed reference sits $0.56$ patches from the other channels on average while they sit $0.97$ from each other. We therefore draw the reference uniformly at random for each sample. This leaves the overall mean displacement unchanged at $0.87$ but equalises the per-channel error across the run.

\begin{figure}[t]
  \centering
  \includegraphics[width=0.68\linewidth]{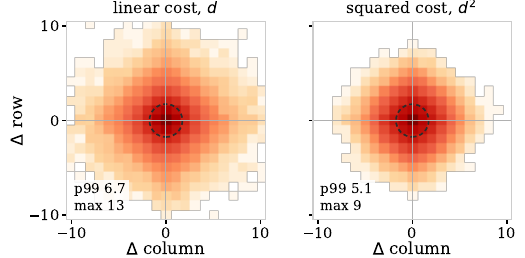}
  \caption{Displacement between two query channels with linear vs.\ squared assignment cost, on a $14 \times 14$ patch grid with $\nchan = 8$ channels. Colour gives the log density of the row and column offset between partnered patches at a mask ratio of $0.6$, and the dashed circle marks is the mean. The two costs carry the same mean and differ only in how far the tail reaches.}
  \label{fig:cost-density}
\end{figure}

The distance in \cref{eq:assign} is L2 as opposed to L1, as we find that a linear cost is more tolerant to longer pairings. \Cref{fig:cost-density} compares the displacements the two costs leave between query channels. They are indistinguishable at the centre and exhibit the same mean. However, the difference is in the tail, where the linear cost leaves $0.36\%$ of pairs more than $8$ patches apart and stretches as far as $13$, against $0.01\%$ and a worst case of $9$ under the squared cost. Squaring therefore minimises the distant pairings.
\section{Experiments}
\label{sec:experiments}

\paragraph*{Datasets and tasks.}

We evaluate on six benchmarks, three of which are classification and three segmentation. CHAMMI \citep{chen2023chammi} and JUMP-CP \citep{chandrasekaran2023jump} are in the IF microscopy domain. CHAMMI is a micropscopy imaging benchmark, which draws single-cell crops from  WTC-11 \citep{viana2023integrated} at $3$ channels, HPA \citep{thul2017subcellular} at $4$, and Cell Painting \citep{bray2016cell} at $5$, and trains one encoder jointly over all three. JUMP-CP is a drug perturbation classificaiton task and contains $8$ channels, $5$ fluorescent and $3$ brightfield. BigEarthNet \citep{sumbul2019bigearthnet} is multi-label land cover classification over $12$ Sentinel-2 bands. 38-Cloud \citep{mohajerani2018cloud} is cloud segmentation in satellite imagery at $4$ channels, and IMC-Breast \citep{jackson2020single} and IMC-Pancreas \citep{damond2019map} are segmentation in imaging mass cytometry at $40$ and $37$ channels. 

\paragraph*{Baseline comparisons.}
We compare against ChannelViT \citep{bao2024channelvit}, DiChaViT \citep{pham2024dichavit} and ChA-MAEViT \citep{pham2025chamaevit}. We consider ChannelViT to be the baseline MC-ViT which consists of channel sampling and channel embeddings. We do not include ChAdaViT \citep{bourriez2024chadavit} in our experiments as it is well established that it is channel sampling that largely contributes to acceptable partial channel metrics on MCI benchmarks. DiChaViT builds on ChannelViT with diverse losses, and ChA-MAEViT builds on DiChaViT with masked image modelling \citep{he2022mae}. We build DCViTv2 on top of ChA-MAEViT for our experiments. Original DCViT \citep{marikkar2026dcvit} has shown to significantly outperform existing methods on non-MIM baselines, we therefore do not waste compute resources to re-run DCViTv2 against those baselines. 

\paragraph*{Model training.}
For all experiments, we use a ViT-S/16 backbone \citep{dosovitskiy2021image}. For all methods if not previously reported in literature, we carry out a light LR sweep on the validation set, between $10^{-4}$ and $10^{-3}$. The rest of the hyperparameters are set from the respective repositories of existing works. The effective batch size is $64$ on CHAMMI and JUMP-CP, and $128$ on the rest.

Where masked image modelling is used for classification, the mask ratio is fixed to the recommended values in ChA-MAEViT \citep{pham2025chamaevit}. For segmentation tasks, we set it at $0.6$. We find this value balances a pretext task that is hard enough while being solvable. For the rest of the hyperparameters, we follow the exact settings as ChA-MAEViT. We run our experiments on a cluster of RTX 3090 GPUs, with the effective batch size kept same as above whenever multi-GPU training is carried out. 

\paragraph*{Evaluation protocol.}

On CHAMMI we use the evaluation code of \citet{chen2023chammi}, in which a nearest-neighbour classifier predicts a macro-averaged F1 per task, and report the mean over the WTC-11 and HPA tasks, following \citet{pham2024dichavit}. On JUMP-CP we report top-1 accuracy over the full channel set and over a reduced set. On BigEarthNet we report mean average precision over its multi-label classes across the twelve Sentinel-2 bands. On 38-Cloud we use the official split, where the test scenes are held out from training. IMC-Breast is a binary tumour against stroma segmentation, and is scored by IoU on the tumour class. IMC-Pancreas is scored by macro IoU across its four cell classes, namely islet, exocrine, immune and other, against a background of tissue carrying no cell.

Every benchmark other than CHAMMI and 38-Cloud is also evaluated under a reduced channel set, which the tables report as partial. The model is trained once on the full set, and channels are withheld only at inference. On JUMP-CP we keep the five fluorescence stains, Mito, AGP, RNA, ER and DNA, and drop the three brightfield channels. This is the partial panel of \citet{bao2024channelvit} and \citet{pham2025chamaevit}. On BigEarthNet we keep the four bands acquired natively at 10\,m, blue, green, red and near-infrared, and drop the eight bands acquired at 20\,m and 60\,m. On IMC-Breast we drop the eight markers that report epithelium, namely cytokeratins 3, 5, 6 and 35, keratin-14, EpCAM, E-cadherin and panCK, while keeping HER2 and $\beta$-catenin. On IMC-Pancreas we drop the five immune markers CD68, MPO, CD20, CD3e and CD45, and score IoU on the immune class. In both IMC panels the markers withheld are the ones that define the class being scored.

\paragraph*{\method{} settings.}
The set $\mixset$ is determined by the task. For segmentation we instrument the four blocks tapped by the segmentation head of existing pixel decoders \citep{ranftl2021vision,pham2025chamaevit}, so that every feature map the head fuses has just been mixed across channels. For classification the read-out is a single pooled vector. However, the global content summarised in this pooled vector is formed in the later blocks \citep{amir2021deep}. Therefore, we employ channel attention in the last four layers at stride two. That is, for a 11-layer encoder, we set $\mixset=\{4,6,8,10\}$. 

In DC-ViT the coefficient $\amix_\lay$ is a single learnable scalar per block in $\mixset$, whose starting value is chosen by hand. Across datasets the value it converges to rises with the number of channels, so rather than fixing one starting value we read it from the input. Since \cref{eq:dsa} weighs a channel axis of length $\nchan$ against a spatial axis of length $\ntok$, we scale by the relative size of the two, setting $\amix_\lay = \amixhat_\lay \cdot \ln (\nchan) / \left( \ln (\nchan) + \ln (\ntok) \right)$, where $\amixhat_\lay$ is the learnable parameter, initialised at one.  $\amix_\lay$ ranges from $0.172$ at three channels to $0.410$ at forty on our benchmarks.

\section{Results}
\label{sec:results}

\subsection{MCI benchmarks}
\label{sec:main-results}
\Cref{tab:main-cls} compares \method{} against the MC-ViT baselines on classification. Every arm shares the training protocol. \method{} is the strongest arm in every column, across fluorescence microscopy and satellite imaging, and it keeps the lead when channels are withheld at inference.

\Cref{tab:main-seg} reports the same comparison on segmentation. \method{} again leads every column, on cloud masking and on the two imaging mass cytometry panels. The lead is again kept when markers are withheld at inference.

\begin{table}[t]
  \centering
  \caption{Classification benchmarks. CHAMMI is the mean macro-F1 (\%) over the WTC-11 and HPA OOD tasks. JUMP-CP (top-1 accuracy \%) and BigEarthNet (mAP \%) are reported over the full channel set and over the reduced set held out at inference.}
  \label{tab:main-cls}
  \begin{tabular}{l c cc cc}
    \toprule
    & CHAMMI & \multicolumn{2}{c}{JUMP-CP} & \multicolumn{2}{c}{BigEarthNet} \\
    \cmidrule(lr){2-2} \cmidrule(lr){3-4} \cmidrule(lr){5-6}
    Method &  & full & partial & full & partial \\
    $\nchan$ & 3--5 & 8 & 5 & 12 & 4 \\
    \midrule
    ChannelViT & 64.97 & 67.51 & 56.49 & 64.71 & 59.50 \\
    DiChaViT   & 69.77 & 69.19 & 57.98 & 64.48 & 59.69 \\
    ChA-MAEViT & 74.63 & 90.73 & 68.05 & 70.47 & 67.54 \\
    \midrule
    \method{}  & \best{78.06} & \best{95.11} & \best{74.05} & \best{70.86} & \best{68.46} \\
    \bottomrule
  \end{tabular}
\end{table}

\begin{table}[t]
  \centering
  \caption{Segmentation benchmarks. 38-Cloud is scored on the official split, where the test scenes are held out from training. IMC-Breast is IoU on the tumour class and IMC-Pancreas is macro IoU over its four cell classes. The partial columns withhold markers at inference. Higher is better in every column.}
  \label{tab:main-seg}
  \begin{tabular}{l c cc cc}
    \toprule
    & 38-Cloud & \multicolumn{2}{c}{IMC-Breast} & \multicolumn{2}{c}{IMC-Pancreas} \\
    \cmidrule(lr){2-2} \cmidrule(lr){3-4} \cmidrule(lr){5-6}
    Method &  & full & partial & full & partial \\
    $\nchan$ & 4 & 40 & 32 & 37 & 32 \\
    \midrule
    ChannelViT & 63.69 & 82.36 & 77.82 & 54.08 & 48.37 \\
    DiChaViT   & 66.14 & 82.56 & 77.89 & 54.64 & 48.73 \\
    ChA-MAEViT & 69.86 & 82.90 & 78.19 & 55.30 & 49.34 \\
    \midrule
    \method{}  & \best{72.96} & \best{86.27} & \best{80.36} & \best{58.01} & \best{51.45} \\
    \bottomrule
  \end{tabular}
\end{table}

\subsection{Analysis}
\label{sec:analysis}

\paragraph{Token correspondence.}
\Cref{eq:assign} pairs the retained patches of two channels by solving a linear assignment on the grid. To ask how much of the benefit comes from solving that assignment well, we vary the rule that establishes correspondence while holding the rest of the model fixed. Three rules are compared. The first is the naive method, where retained patches of each channel are sorted along the absolute index, where the absolute index is based on a row-major traversal along the square grid. The sorted patches are then partnered with the patches of other channels at the same newly sorted index. The second is similiar rank matching as above, but the index sorting is based on a generalised Hilbert traversal along the grid \citep{hilbert1891curve,zhang2006pseudohilbert}. This reduces the $x-y$ asymmetry in mean correspondence error which exists in row-major traversal. The third is the learned assignment \cref{eq:assign} which minimises the total distance between partners while preserving a 1-1 mapping. The first two require no solver as the traversals are fixed. 

\begin{figure}[t]
  \begin{minipage}[t]{0.48\linewidth}
    \centering
    \includegraphics[width=\linewidth]{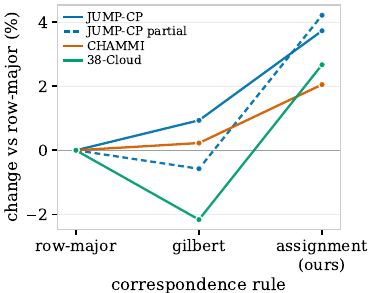}
    \caption{Each correspondence rule against the change it produces, relative to row traversal, in the metric reported for each benchmark. The mean displacement it leaves is $4.0$ to $7.8$ patches under row traversal, $1.8$ to $3.6$ under the Hilbert traversal and $0.9$ to $1.9$ under the assignment.}
    \label{fig:correspondence}
  \end{minipage}\hfill
  \begin{minipage}[t]{0.48\linewidth}
    \centering
    \includegraphics[width=\linewidth]{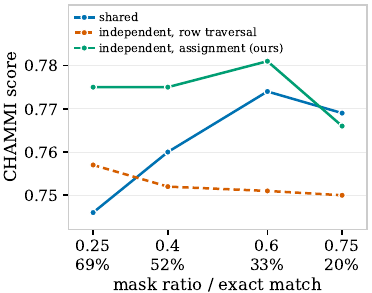}
    \caption{CHAMMI under shared and independent masking, across mask ratios. Beneath each ratio is the share of compared patches the assignment aligns exactly, measured from the masks alone.}
    \label{fig:masking}
  \end{minipage}
\end{figure}

\cref{fig:correspondence} shows the variation in downstream performance for JUMP-CP, 38-Cloud and CHAMMI, under various degrees of correspondence error induced by the above methods. The linear assignment leaves the lowest error and gives the highest performance on all three. The two traversals, however, provide similar results, although the Hilbert traversal roughly halves the mean error of the row-major one. The mean error alone therefore does not account for the difference.

What the assignment changes is which patches are partnered, rather than how far apart they are. A traversal sorts each channel on its own, so a patch retained in two channels is placed at the same index only when the number of patches before it agrees in both. The assignment partners such a patch with itself, as a distance of zero cannot be improved upon. On JUMP-CP, the share of compared patches that are the same patch is $10.8\%$ under the row-major traversal and $10.9\%$ under the Hilbert traversal, against $61.9\%$ under the assignment. On 38-Cloud the three are $3.1\%$, $3.0\%$ and $29.7\%$. Perfectly matched partners are therefore high-value patches that have a strong effect on downstream performance.

\paragraph{Robustness to the mask ratio.}
Shared masking gives every channel the same retained patches, so every partner is the same patch and correspondence is exact by construction. On the argument above it should therefore be the strongest of the three settings. \Cref{fig:masking} shows that it is not. On CHAMMI it trails independent masking with the assignment at every ratio below $0.75$, by $0.029$ at a ratio of $0.25$ and by $0.015$ at $0.4$, despite aligning $100\%$ of partners against $69\%$ and $52\%$.

Here, exact correspondence is not the only thing that matters. Shared masking obtains it by exposing the same patches in every channel, and in doing so gives up the mask diversity that masked multi-channel training relies on. The assignment, on the other hand, keeps that diversity while recovering most of the correspondence. \cref{fig:masking} shows that assignment stays between $0.766$ and $0.781$ across the range, whereas shared masking spans $0.746$ to $0.774$ and only reaches it at the heaviest ratios. Independent masking with the row-major traversal is flat as well but sits lowest throughout, hence our proposed assignment is what carries the difference.

The share of exact-match pairs accounts for the assignment's shrinking lead over shared masking. Heavier masking leaves fewer patches retained in both channels, so the share the assignment can align exactly falls, from $69\%$ at a ratio of $0.25$ to $33\%$ at $0.6$ and $20\%$ at $0.75$ on a three-channel input. The lead narrows with it, from $0.029$ to $0.007$, and the two are level at $0.75$. Shared masking is therefore competitive only at the heaviest ratios, and the strongest result of the three arms remains independent masking with the assignment, at a ratio of $0.6$.
\section{Conclusion}
\label{sec:conclusion}

In this work, we address two questions left open by DC-ViT. The first is whether decoupled attention extends to dense prediction, and whether it holds for datasets with high channel counts such as imaging mass cytometry. The second is how decoupled attention can be trained with masked image modelling, where each channel is masked independently and the visible tokens no longer align across channels. We show that token correspondence can be recovered by solving a linear assignment between the retained patches of each channel, and that scoring each tuple against a single reference channel reduces an otherwise intractable joint problem to $\nchan - 1$ assignments, each of which is solved exactly. This allows decoupled attention to be combined with masked multi-channel training in its standard configuration, rather than the restricted one it was previously limited to. Extensive experiments across six MCI benchmarks spanning fluorescence microscopy, imaging mass cytometry and satellite imaging show that \method{} consistently outperforms existing MC-ViT methods on both classification and segmentation. Our analysis further shows that the improvement is driven not by the average distance between partnered patches, but by the share of partners that are the same patch, and that recovering correspondence under independent masking is more effective than restricting every channel to a shared mask. Overall, this work shows that explicit token correspondence is what allows structured channel interaction to be retained under masked training, and provides a formulation of decoupled attention that applies across MCI datasets and tasks.

\backmatter

\end{document}